\documentclass{elsarticle}

\makeatletter
\def\ps@pprintTitle{%
	\let\@oddhead\@empty
	\let\@evenhead\@empty
	\def\@oddfoot{}%
	\let\@evenfoot\@oddfoot}
\makeatother

\usepackage{lineno,hyperref}
\modulolinenumbers[5]

\usepackage{cprotect}
\usepackage[export]{adjustbox}

\renewcommand{\arraystretch}{1.1} 

\newcommand{\charite}{Charit\'{e}}

\begin{document}

\begin{frontmatter}

\title{A Case Study on Pros and Cons of Regular Expression Detection and Dependency Parsing for \\Negation Extraction from German Medical Documents. Technical Report}

\author{Hans-J\"{u}rgen Profitlich\corref{mycorrespondingauthor}}
\ead{profitlich@dfki.de}

\author{Daniel Sonntag}
\ead{sonntag@dfki.de}

\address{German Research Center for Artificial Intelligence (DFKI)\\
		Saarbr\"{u}cken, Germany\\
		and Oldenburg University}

\begin{abstract}
We describe our work on information extraction in medical documents written in German, especially detecting negations using an architecture based on the UIMA pipeline. Based on our previous work on software modules to cover medical concepts like diagnoses, examinations, etc. we employ  a version of the NegEx regular expression algorithm with a large set of triggers as a baseline. We show how a significantly smaller trigger set is sufficient to achieve similar results, in order to reduce adaptation times to new text types. We elaborate on the question whether dependency parsing (based on the Stanford CoreNLP model) is a good alternative and describe the potentials and shortcomings of both approaches.
\end{abstract}

\begin{keyword}
information extraction (IE); negation detection, regular expression detection, natural language processing; dependency parsing; electronic health record (EHR)
\end{keyword}

\end{frontmatter}

\section{Introduction}

As medical records may cover a very long history of diseases (up to several decades) and include a vast number of diagnoses, symptoms, results, medications, and laboratory values, we could highly benefit from advanced search capabilities in clinical information systems to allow for the retrieval of relevant data. The first of several steps to process such data is information extraction. The fact that medical texts often contain descriptions of examinations and therapies leads to frequently occurring negated expressions which play a central role in understanding a patient's medical history. For example, a biopsy may include various specific tests and the results state whether a test was positive or negative (``edema, no metastases'').

In comparison to many other text types, electronic health reports, discharge or transfer letters and other kinds of medical reports are often written in a rather telegraphic style with compacted information. In addition, the detection of negation scope in German language is more difficult than in other languages, such as English (cf. for example \cite{Wiegand:2010:SRN:1858959.1858970}). Similar to English, German negations can be directly bound to a word as prefix like in ``unauff\"allig'' (unremarkable) or as suffix like in ``schmerzlos'' (painless). However, German texts differ in various aspects from English texts. For example, German is a richly inflected language (e.g., ``no'' can be translated as ``kein'', ``keine'', ``keiner'' etc.). Furthermore, German includes discontinuous or surrounding triggers, such as ``weder ... noch ...'' (neither ... nor ...).  Another problem is that in most cases these triggers can be reduced and use a different word stem or preposition, like in case of ``wies ... zur\"uck'' (rejected) and ``wies ... nach'' (verify), which can easily lead to too general trigger expressions when using for example regular expressions. Triggers may also  both precede and follow the negated expression. 

This work describes a software system realized within a large medical project about testing AI components (pAItient---developing, testing and evidence based evaluation of clinical value). The task is to extract medical information from unstructured German texts like findings or dismissal letters. We start by implementing an information extraction module  to segment the sentences and cover mentioned medical concepts like diagnoses, disorders, or examinations based on some dictionary annotators. 

To recognize negations we integrate and test an implementation of the NegEx algorithm \cite{DBLP:journals/jbi/ChapmanBHCB01}
which leads to a good baseline, but also reveals some shortcomings described in this paper. We tested and compared our system to the system described in \cite{DBLP:conf/coling/CotikRXUBS16}\footnote{We want to thank the authors for providing the data for this study.}.  After studying the errors we were able to establish some improvements by optimizing the code and adjusting the configuration. As the configuration of NegEx comprises mainly the trigger set and the size of the window, we conducted  experiments regarding an easier set of triggers  and different scopes.
As a result, we describe how the trigger set can be reduced to a small number of patterns (56 instead of 506 triggers) without decreasing quality, to make such approaches more practical and applicable in changing text extraction contexts. 

In the rest of this study we show, first, that the trigger set can be reduced to a very small number of patterns without decreasing the classification quality (on our real world data set). Regular expressions as used in NegEX rely solely on surface text, and thus are limited when attempting to capture complex syntactic constructions such as long noun phrases. Second, to overcome the inherent shortcomings of this methods processing the surface text only, we add several natural language processing (NLP) modules including a dependency parser, which is often mentioned in the literature (e.g.,  in \cite{DBLP:conf/coling/CotikRXUBS16}) when dealing with these problems. We evaluate several NLP modules including the Stanford CoreNLP dependency parser. In this case study we describe details of our implementation and the pros and cons of both approaches (regular expression detection vs. dependency parsing). For example, we show why dependency parsing is neither a good alternative nor a satisfactory complement for our type of medical documents. Deep syntactic processing turned out to be helpful in some cases, but implementing it is very time-consuming and entails other risks and problems. We then share some insights about the structure of trigger sets and the scope of triggers based on regular expressions.

\section{Related Work and Background}

Clinical information extraction from patient records is still underrepresented and underdeveloped in clinical settings. The identification of semantic relations, such as substance A treats disease B, remains a non-trivial task \cite{sonntag03}.  A special trend becomes apparent, the need for ontology modeling of medical terminology and corresponding information extraction results \cite{SonntagWBZ09}. Because of enormous annotation costs, mainly unsupervised methods are being used \cite{Alicante2016}. Only recently, new text mining approaches on medical literature have been proposed, for extracting adverse drug events from text \citep{ad15} or automatic symptom extraction on rare diseases \citep{sym15} for example. 

Other research has been dedicated to clinical negation detection together with the detection of pathological entities in German texts. Bretschneider et al. \cite{DBLP:conf/cbms/BretschneiderZH17} classify sentences containing pathological and non-pathological findings in German radiology reports. Their approach uses  syntacto-semantic parsing methods. Gros and Stede \cite{grosstede2013} present Negtopus, a system that identifies negations and their scope in medical diagnoses written in German and in English.

Even though many NLP applications have been developed recently that successfully extract implicit facts mentioned in medical reports, discriminating between positive, negative, and uncertain findings remains challenging \cite{ DBLP:conf/lrec/Morante10, DBLP:journals/jdi/WuDKR11, DBLP:conf/amia/WuMMCCHHC13, DBLP:conf/naacl/GkotsisVODLD16}.

Rule-based systems rely on negation keywords and rules to determine the negation scope \cite{DBLP:journals/jamia/FriedmanHSL99}. 
There are two main approaches: 1) directly on surface text or 2) based on examining the structures resulting from natural language parsing.
Research on processing negation has been carried out mostly on clinical reports, and has focused on detecting whether a medical term is negated or not. Various approaches have been published for detecting negations and speculations in English medical reports.
A widely used and popular tool is NegEx \cite{DBLP:journals/jbi/ChapmanBHCB01,DBLP:conf/medinfo/ChapmanHVKSCCTMD13, DBLP:conf/medinfo/MitchellBBCGGHL04}. 
The approach uses a simple algorithm based on a list of triggers (or simple regular expressions) that indicate negation and speculation. A hard-coded window of words preceding or following a trigger is used to determine the scope of a negation or a speculation (uncertainties or equivocal findings like ``cannot be excluded'').
The original NegEx algorithm with hard-coded scopes was extended to ConText by Harkema et al. \cite{DBLP:journals/jbi/HarkemaDTC09} by extending the scope of a negating term up to the end of a sentence or a termination term (called ``conjunction'') like ``but''. Additionally, there are left-looking trigger terms that scope to the beginning of a sentence. Obviously, the results still depend very much on the used window size and the type of the texts.

NegEx has also been adapted to French, Swedish, Spanish, German and other languages with good results \cite{DBLP:journals/biomedsem/Skeppstedt11,DBLP:conf/ihi/DelegerG12,DBLP:conf/bionlp/CotikSVR16,DBLP:conf/brain/CostumeroLGMM14,DBLP:journals/bmcbi/AfzalPKSSK14}. 
Approaches regarding the negation scope delimitation has been done by \cite{DBLP:conf/nodalida/TanushiDDKSV13} for texts in Swedish.
Beside NegEx  a wide range of other methods exist, for example based on syntactic techniques \cite{DBLP:journals/jamia/MutalikDN01,DBLP:journals/jamia/HuangL07,DBLP:conf/bionlp/CotikSVR16} or machine learning techniques \cite{DBLP:journals/jamia/UzunerZS09}. The i2b2 challenges also include contests for assertion classification which show very good results (\cite{DBLP:journals/jamia/BruijnCKMZ11, DBLP:journals/jbi/BejanVXY13}).

However, these systems have two major drawbacks: 1) they rely on the lexical structure of sentences and 2) the scope of a negation is somehow arbitrarily fixed, i.e., they cannot cope with the problem of long distance information between negation keywords and the target. For example, a pattern covering an expression like ``no signs of an infection'' does not match the phrase ``no detectable signs of a new infection''.
A number of systems tried to enhance the results of these approaches based on surface text and to overcome their shortcomings by additionally analyzing the syntactic structure of sentences, most notably dependency graph, the associated set of directed binary grammatical relations that hold among the words. Relations among the words are illustrated in our figures and medical example sentences with directed, labeled arcs from heads to dependents. We call this a typed dependency structure in computational linguistics.\footnote{Please have a look at Daniel Jurafsky`s  and James H. Martin`s informative chapter in Speech and Language Processing (forthcoming), see \url{https://web.stanford.edu/~jurafsky/slp3/14.pdf}}.

Sohn et al \cite{Sohn2012DependencyPN} used regular expressions on the dependency path and Mehrabi et al \cite{DBLP:conf/medinfo/MehrabiSWBKKDATP13, DBLP:journals/jbi/MehrabiKSRSKBDS15} used dependency patterns as a post-processing step after NegEx to remove false positives. A similar approach can be found by Savova et al \cite{DBLP:journals/jamia/SavovaMOZSSC10} who realized the text analysis and knowledge extraction system cTAKES. Ballesteros et al \cite{DBLP:conf/cicling/BallesterosFDHG12} also tried to infer the scope of negations by using a dependency parser. Huang and Lowe \cite{DBLP:journals/jamia/HuangL07} proposed a classification scheme of negations based on syntactic categories and patterns in order to locate negated concepts, regardless of their distance from the negation cue. Weng et al \cite{DBLP:journals/corr/abs-2003-00353} are using NegEx and the parsing tree, but also concept identification using metathesaurus, and parsing tree pruning to capture important parts for negation. \cite{negbio} and \cite{DBLP:conf/aaai/IrvinRKYCCMHBSS19} also use the dependency parsing for negation detection and propose a relatively comprehensive rule set (but only for English).

\section{Pipeline Implementation}

In our system, information extraction is performed using  UIMA (Unstructured information management architecture) \cite{OASIS:UIMA:2009}, a framework of software systems for analyzing large volumes of unstructured information. UIMA is a widely used architecture in NLP systems because of its support for modular combination of components, reuse of code, and interoperability. We employ the pipeline to extract medical data about diagnoses, symptoms, medications, and laboratory values from the documents. The central data structure is a so-called CAS object (common analysis structure), which is guided through an analysis pipeline. The modules of this pipeline, called analysis engines (AE), can access annotations of previous modules and enrich the CAS with additional annotations. 

We use implementations of open source libraries where applicable and appropriate, mainly the DKPro Core \cite{eckartdecastilho-gurevych:2014:OIAF4HLT}\footnote{\url{https://dkpro.github.io/dkpro-core/}}, a large collection of open source software components for NLP based on the Apache UIMA framework. Several of these DKPro analysis engines are based on the Stanford NLP  package\footnote{\url{https://stanfordnlp.github.io/CoreNLP/}}\cite{manning-EtAl:2014:P14-5}. 
In some cases we adapt and extend the DKPro code due to our requirements.

\begin{figure}[t]
	\centering
	\includegraphics[width=0.8\textwidth]{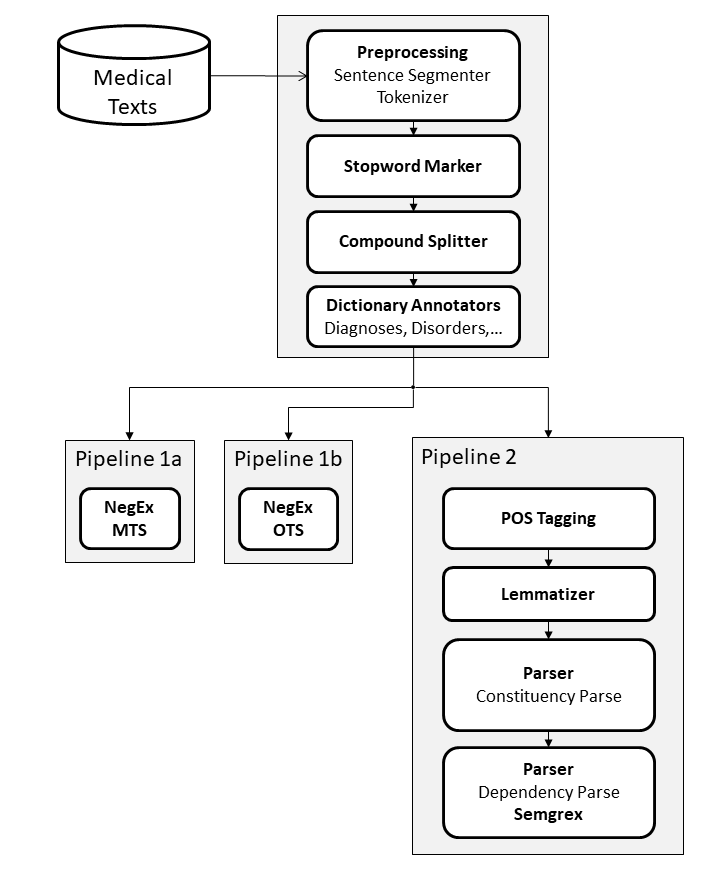}
	\caption{UIMA Pipeline showing the common starting block of analysis engines up to the dictionary annotators and the following negation detection variants (Pipeline 1a, 1b and 2).}
	\label{fig:uimapipeline}
\end{figure}

Our UIMA pipeline starts with some basic preprocessing steps to annotate sentences, tokens, and stop words (own implementations if not marked otherwise), as below:
\begin{itemize}
	\item CoreNlpSegmenter: (DKPro) the Stanford NLP Segmenter to split the input text into sentences and tokens;
	\item PatternBasedSentenceSegmenter, PatternBasedTokenSegmenter: (DKPro) used to split sentences and tokens further by regex-patterns;
	\item StopwordMarker: annotates stopwords like ``is'', or ``and'', i.e., words that can be ignored for our purposes. It is based on a manually assembled list of words.
\end{itemize}

Additionally there are further modules to decompose German compounds and to annotate medical concepts like diagnosis, disorders, etc.
\begin{itemize}
	\item CompoundSplitter: splits German compounds into its components (based on the code of D. Naber\footnote{\url{https://github.com/danielnaber/jwordsplitter}});
	\item DictionaryAnnotator(s): adaption of a DKPro implementation of a dictionary annotator; for diagnoses, disorders, symptoms, examinations, etc.
\end{itemize}

Modules needed for the additional dependency parsing:
\begin{itemize}
	\item CoreNlpPosTagger: the Stanford NLP POS (part-of-speech) tagger;
	\item IxaLemmatizer: (DKPro) IXA Lemmatizer (the Stanford NLP Lemmatizer does not support German);
	\item CoreNlpParser: Stanford NLP parser;
	\item CoreNlpDependencyParser: adaptation of the Stanford NLP dependency tagger, integrates a module to execute Semgrex patterns on the semantic graph.
\end{itemize}

\subsection{Pipelines 1a and 1b: Negation Extraction}

NegEx  is a widely used algorithm that utilizes regular expressions (called triggers) to detect negations. Triggers come in different versions depending on their position relative to the negated term: the main types are pre-negations, post-negations, and pseudo-negations. Pseudo-negations are used to avoid annotating a negation when a pre-negation trigger refers to a word that is not a medical finding, like in ``keine \"Anderung ...'' (no change) or ``kein Anstieg ...'' (no increase). Conjunctions are used to terminate the scope of a trigger.
We aimed at comparing our results to the ones described in Cotik et al. \cite{DBLP:conf/coling/CotikRXUBS16} (referred as the MACSS project) which used a Python implementation of NegEx. Their implementation does not use regular expression triggers for German texts but plain strings. They translated the original trigger set of Chapman et al. and  expanded it which results in a set of 506 ``MACSS'' triggers\footnote{This trigger set can be found here: \url{http://macss.dfki.de/negex/negex_trigger_german_biotxtm_2016.txt}}. 
These MACSS triggers use long expressions covering variations of wordings like ``does not apply to the patient'', ``did not apply to him'', ``did not apply to her'', which makes them hard to handle and maintain. We tested whether it's sufficient to use the actual negation parts only (``not'' in these cases) without losing precision by the resulting greater distance between trigger and negated term. This results in a trigger set of only 56 items, a reduction by 88\% (see also section \ref{sec:comppip}). 

Our own implementation of NegEx is based on the Java software of the Google code archive\footnote{\url{https://code.google.com/archive/p/negex/}}. It had to be extended, as it originally just returned the negated text part but not the negating expression which is necessary to validate the results. 

In contrast to the MACSS implementation we did not include the annotation of uncertainties (like ``not sure'' for example), as medical statements and wordings mostly avoid explicit unique expressions anyway. Instead, often phrases like ``cannot be ruled out'', ``no signs of'' are used, which always leave some remaining uncertainty. Furthermore the MACSS trigger set contains only 13 triggers for uncertain negations and the only one that occurred in the example texts was the question mark.

The  analysis engine that wraps the NegEx algorithm can be applied as a trailing engine to the main pipeline described above (see pipelines 1a/b in figure \ref{fig:uimapipeline}) and can be called with both the MACSS trigger set called ``MTS'' (pipeline 1a) or our own set (``OTS'', pipeline 1b).

\subsection{Pipeline 2: Dependency Parsing}

As a second extension of the pipeline (see pipeline 2 in figure \ref{fig:uimapipeline}) we integrated some natural language modules from the Stanford NLP project (or rather their DKPro implementation): part-of-speech tagger, lemmatizer, parser, and dependency parser.

The Stanford NLP framework  includes a mechanism to match patterns against the dependency trees of sentences, called Semgrex\footnote{There are similar modules to match sequences of tokens (TokenRegex) or partial trees (Tregex)}.
We implemented a module based on the Semgrex mechanism \cite{DBLP:conf/acl/ChambersCGHKMMR07}\footnote{\url{https://nlp.stanford.edu/software/tregex.html}} to match negation triggers and the corresponding negated tokens from the dependency graphs. 
The Semgrex mechanism works directly on the Stanford type system which made it necessary to integrate our module  into the DKPro implementation of the dependency parser. A separate analysis engine would have been preferred but the trees are not stored in the data structure of the pipeline and there is no implementation available that re-constructs a semantic graph out of the annotations in German. 
The module uses a set of patterns defining various negations phrases.

A simple Semgrex pattern defines two nodes and some relation between them\footnote{A more detailed description can be found here: \url{https://nlp.stanford.edu/nlp/javadoc/javanlp/edu/stanford/nlp/semgraph/semgrex/SemgrexPattern.html}}.
A node is represented by a set of attributes and their values contained by curly braces: \verb|{attr1:value1;attr2:value2;...}|, for example \verb|{pos:/NN/}| refers to a node whose part-of-speech annotation is ``NN''.
Attributes must be plain strings, common used attributes are ``pos'' (part-of-speech), ``lemma'', or ``word'' (the word itself in lowercase characters). Values can be strings or regular expressions blocked off by ``/''.
\verb|{}| represents any node in the graph. 
A node can be given a name by appending ``=\textit{name}'' to the node specification which allows us to access specific nodes after a match. Relations are defined by a symbol representing the direction of the relationship and a string or regular expression representing its value, e.g. ``$>$ /nmod/''. The symbols ``$<$'' and ``$>$'' indicate the direction of the relation, and ``$>>$'' stands for a relation chain.
We defined two types of fixed names: ``gov'' depicts the governor of a negation (the negating expression) and names starting with ``dep'' link to all dependents (the negated expressions). These names are used by the analysis engine to annotate the named tokens when a pattern matches.

As we realized two methods to detect negations (NegEx and dependency parsing), the patterns can be used for different purposes: 
1) detect negations that NegEx has missed or 
2) correct annotations falsely marked as negations by NegEx.

This leads to (at least) two types of patterns:
\begin{itemize}
	\item patterns to mark negated terms: confirmation or supplement of annotations by NegEx
	\item patterns to mark positive terms: correction of false annotations by NegEx, e.g. ``nicht ausgeschlossen'' (not excluded)
\end{itemize}
Patterns are read from resource files where they are marked with the according type flags. Some examples of Semgrex patterns and their meaning are given in table \ref{tab:patterns}.

\begin{table}
	\centering\small
	\begin{tabular}{|l|l|}
		\hline 
		Pattern & \verb§{lemma:/frei/}=gov > /nmod:von/ {}=dep§\\
		Description & A node with lemma $=$ `frei' (\textit{free}, referenced as `gov') is in relation\\&  of type `nmod:von'  to another node (referenced as `dep')\\
		Example & \textit{Patient frei von Schmerzen} (patient free of pain) \\
		\hline 
		\hline
		Pattern & \verb§!{lemma:/anzeichen|hinweis/}=dep > /neg/ {}=gov§\\
		Description & A node whose lemma is not `anzeichen' or `hinweis' (\textit{signs/hints}, \\&  referenced as `dep') is in relation of type `neg' to another node\\&  (referenced as `gov')\\
		Example & \textit{Keine Infektion erkennbar} (no infection recognizable)\\&
		but not matching \textit{Keine Anzeichen einer Besserung}\\& (no signs of improvement) \\
		\hline
		\hline
		Pattern & \verb§{pos:/NN/}=dep < /.*subj.*/§\\ &\verb§ ({word:/ausgeschlossen|auszuschlie.*en/}=gov§\\
		&\verb§ !>> /neg/ {})§\\
		Description & A node whose `pos' is not `NN' (referenced as `dep') is the target of a \\& relation of type `subj' (including variants like `nsubj') to another node\\&  (referenced as `gov') whose word (text) is  `ausgeschlossen' (\textit{excluded})  \\& and there is no relation chain containing a relation of type `neg'\\& to another node. \\
		& The part `\verb|!>> /neg/ {}|' ensures that  `ausgeschlossen' is not negated.\\
		Example & \textit{Fieber ist ausgeschlossen} (fever is excluded)\\&
		but NOT matching \textit{Fieber ist nicht v\"ollig ausgeschlossen}\\& 
		(fever is  not excluded completely). \\&This would be an example of a \textit{double-negation}. \\
		\hline
	\end{tabular}
	\caption{Some examples of Semgrex patterns: `gov' is the governor (the \textit{negating} term) and `dep' is the dependend (the \textit{negated} term) of the negation. `pos' stands for part-of-speech, `NN' for a noun.}
	\label{tab:patterns}
\end{table}

\subsection{User Interface}

\begin{figure*}
	\centering
	\includegraphics[width=\textwidth, frame]{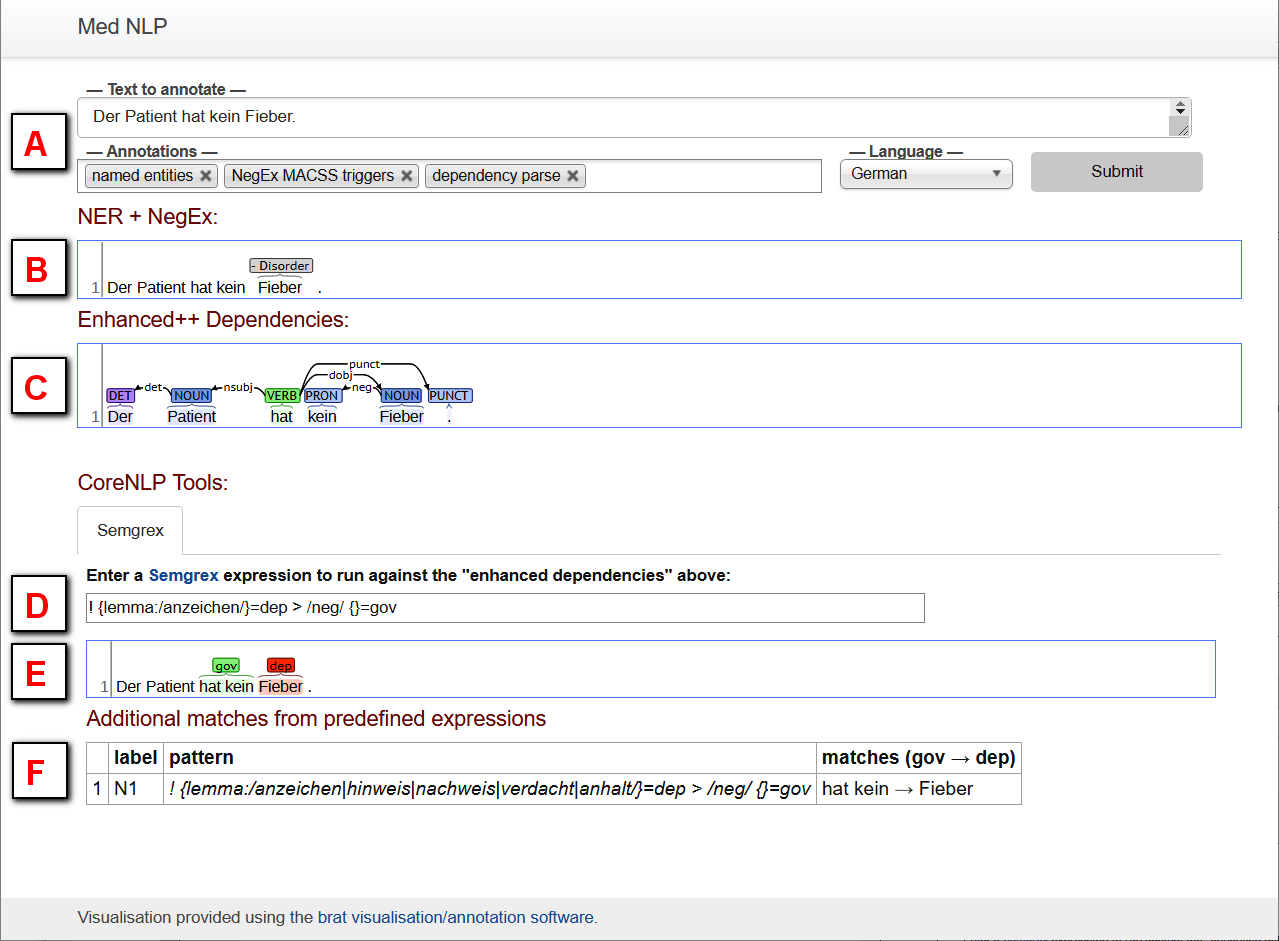}
	\caption{Web-based user interface. A: text input field plus menus to determine language and annotations, 
		B: results of named entity detection: negated terms are marked with a ``-'' sign, 
		C: result of dependency parsing,
		D: input field for an interactive Semgrex pattern,
		E: results of the interactive Semgrex pattern,
		F: additional results of predefined Semgrex patterns}
	\label{fig:Workbench}
\end{figure*}

Based on the brat visualisation software\footnote{\url{http://brat.nlplab.org}} \cite{stenetorp2011supporting} we implemented a web workbench that enables the user to enter both sample sentences and Semgrex patterns, and which presents the results of predefined patterns plus the matches of the interactively entered pattern. This is of great help when specifying and tuning new Semgrex patterns.
Additionally, the web frontend allows us to directly compare the results of the UIMA pipeline, especially the negations found by the NegEx modules with the matches of Semgrex patterns.
Figure \ref{fig:Workbench} shows the different parts of  this interface.

\section{Evaluation Data}

The evaluation data originates from the MACSS project and is the same as described in Cotik et al. \cite{DBLP:conf/coling/CotikRXUBS16}. It has been provided by \charite\ Berlin and
comprises of two types of documents from the nephrology domain: discharge summaries and clinical notes. Both types are written by medical doctors during their daily routine, but have significant differences: the clinical notes are rather short and are written shortly after a visit of a patient. Discharge summaries are written to summarize a stay at the hospital and are more structured. 

It should be mentioned that the original form of the document text was not available for evaluation.
Instead, all documents are provided in a format where each line contains exactly one medical concept, followed by the whole sentence and a trailing tag that can be either ``Affirmed'' or ``Negated''. This is important as what is regarded as a ``sentence'' was determined by an algorithm, which splits a document into sentences according to a specific delimiter patterns like a punctuation mark (.) or a double newline for example. As the medical texts contain enumerations in a tabular like style (without punctuation marks), the resulting sentence annotations are sometimes very long (and often longer than a human reader would tell when parsing the document structure). This is important to note as the NegEx algorithm assigns the scope of negations within whole sentences which is more error-prone when (natural) sentence boundaries are not detected. In Cotik et al.\ the annotations have been made manually by two annotators. Some smaller inconsistencies remained as sometimes in noun phrases like ``acute anemia'' the complete phrase was marked and in other cases just the word ``anemia''.  Additionally, sometimes a correct annotation in a sentence was completely missing. We corrected these inconsistencies in the ground truth documents. Beside this, we ignored annotations originally marked as ``speculated'' as we do not annotate them. (This led to the effect that our results differ a bit from the results originally reported by Cotik et al., the systems however remain comparable in a quantitative evaluation with the corrected dataset.) These documents served as ground truth for our evaluation.\footnote{The data format does not specify the exact positions of the medical concept or the negation term. This means it is often not clear which occurrence of a concept is referred to if the same term occurs more than once in a sentence, or which negation term is related to a specific concept (sentences often contain multiple medical concepts and negation terms).} 


Our ground truth data set with corrected annotations has been split (as also usual in machine learning experiments) in a learning dataset (set 1) and a testing dataset (set 2).   Table \ref{tab:documents} provides an overview of the learning set (called set 1). Table \ref{tab:annotations} shows the number of findings (annotations) that are affirmed and the number of negated terms in the documents serving as training set (after completing the annotation). Here, training means the manual iterative extraction rule generation process for a specific pipeline. For example, set 1 used to improve our NegEx algorithm and to tune the trigger sets (mainly by manually checking the false positives and false negatives and refining NegEx rules). Additional documents (also from the MACSS project) have been made available that were used solely for evaluation (referred to as set 2). Set 2  comprises of 45 discharge summaries and 175 clinical notes (see table \ref{tab:docplus}).  

\begin{table}
		\centering
	\begin{tabular}{|l|r|r|}
		\hline 
		& discharge  & clinical  \\ 	
		&  summaries &  notes \\ 	\hline
		\# documents       &    8 &  175 \\ 	\hline
		total words     & 6221 & 6674 \\ 	\hline
		total sentences & 1076 & 1158 \\ 	\hline
		avg. words per doc. & 777.63 & 38.14 \\ 	\hline	
	\end{tabular}
	\caption{Statistics about the annotated documents (training set 1)}
	\label{tab:documents}
\end{table}

\begin{table} 
		\centering
	\begin{tabular}{|l|r|r|}
		\hline 
		  & discharge  & clinical  \\ 	
		  &  summaries &  notes \\ 	\hline
		
		affirmed & 409 & 284 \\ 	\hline
		negated  & 108 & 338 \\ 	\hline
		findings & 517 & 622\\ 	\hline			
	\end{tabular}\\
	\caption{Number of affirmed and negated terms in set 1}
	\label{tab:annotations}
\end{table}

\begin{table}
	\centering
	\begin{tabular}{|l|r|r|}
		\hline 		 
		  & discharge  & clinical  \\ 	
		  &  summaries &  notes \\ 	\hline
		
		affirmed & 2421 & 261 \\ 	\hline
		negated  &  756 & 306 \\ 	\hline
		findings & 3177 & 567 \\ 	\hline		
		
	\end{tabular}\\ 
	\caption{Statistics about the annotated documents for evaluation (testing set 2)}
	\label{tab:docplus}
\end{table}

\section{Experiments}

Our original pipeline was designed to recognize different types of medical concepts like diagnoses, disorders, or examinations for example. These dictionary annotators are trained based on terminology sources such as ICD-10\footnote{\url{https://icd.who.int/browse10/2016/}}. However, the documents used as a ground truth were annotated (manually by human annotators) based on different sources.
In order to be able to compare the results we adapted the analysis engines of our pipeline: instead of searching for diagnoses, examinations etc. with different annotators we defined one special ``med\_concept'' annotator that uses exactly the set of medical concepts collected from the training set. In future investigations, we will use ontology matching to SNOMED\footnote{\url{https://www.snomed.org}}, the standard used in the pAItient project over the next several years (Protected Artificial Intelligence Innovation Environment for Patient Oriented Digital Health Solutions for developing, testing and evidence based evaluation of clinical value).  

\subsection{Comparison of pipelines 1a and 1b}
\label{sec:comppip}

We performed the evaluation on 1) the MACSS triggers used in Cotik et al\footnote{\url{http://macss.dfki.de/negex/negex_trigger_german_biotxtm_2016.txt}}, called ``MTS'' in pipeline 1a, and 2) our own set of triggers ``OTS'' in pipeline 1b. Table \ref{tab:triggersets} shows the different kinds of triggers in the two trigger sets and their counts.

\begin{table}
	\centering
	\begin{tabular}{|l|r|r|}
		\hline 
		trigger	&  MTS    &  OTS \\ 	\hline
		PRE     &  97 & 17 \\ 	\hline
		POST    &  41 & 11\\ 	\hline
		CONJ    & 242     &  7 \\ 	\hline
		PSEU    & 107     & 21 \\ 	\hline	
		total	& 487	& 56	\\ 	\hline	
	\end{tabular}
	\caption{Comparison of trigger sets}
	\label{tab:triggersets}
\end{table}

Besides the set of triggers the only remaining configuration option of NegEx is the size of the window, i.e., the maximum scope of a negation (given as number of words). As a basic rule of NegEx any negation reaches up to the end/beginning of a sentence or to a conjunction. We evaluated all documents with the following window sizes: no restriction, 5, 4, and 3. The complete results are shown in the following tables (\ref{tab:compNegEx1MTS}, \ref{tab:compNegEx1OTS}, \ref{tab:compNegEx2MTS}, and \ref{tab:compNegEx2OTS}). Table \ref{tab:compNegExF1} shows the F1 scores of all variations (highest values are in bold, lowest values in italics).  For comparison, the F1 values reported in Cotik et al. (which used trigger set 1) has been 0.91 for summaries and 0.96 for notes. 

If the size of the scope remains unlimited, the distance between a negation and a concept can be up to 22 words (for document type notes). The reason for this lies in the telegram like writing style of notes which can lead to long 'sentences' which are sometimes just long listings of findings.

{
	\setlength{\tabcolsep}{5pt} 
	\renewcommand{\arraystretch}{1.2} 
	
	\begin{table*}[t!]
		\centering
		\begin{tabular}{|l||r|r|r|r||r|r|r|r|}
			\hline 		  
			F1 & \multicolumn{4}{|c||}{discharge} &  \multicolumn{4}{|c|}{clinical}  \\ 	
			 & \multicolumn{4}{|c||}{summaries} &  \multicolumn{4}{|c|}{notes} \\\hline
			scope 		& $\infty$  & 	  5 &     4 & 3		&  $\infty$ &     5 &     4 &	 3	\\ \hline  
			MTS set 1	& 0.922 & 0.959 & {\bf 0.962} & {\em 0.878}		& {\em 0.953} & 0.971 & 0.974 & {\bf 0.981}	\\ 	\hline		
			OTS set 1	& 0.909 & {\bf 0.954} & 0.933 & {\em 0.832}		& {\em 0.950} & 0.970 & 0.974 & {\bf 0.981}	\\ 	\hline		
			MTS set 2   & 0.900 & {\bf 0.902} & 0.891 & {\em 0.849}		& {\em 0.929} & 0.966 & 0.966 & {\bf 0.974}	\\ 	\hline		
			OTS set 2	& 0.909 & {\bf 0.909} & 0.897 & {\em 0.854}		& {\em 0.934} & 0.971 & 0.971 & {\bf 0.979}	\\ 	\hline		
			
		\end{tabular}\\ 
		\caption{F1 values for all tests}
		\label{tab:compNegExF1}
	\end{table*}
}

{
	\setlength{\tabcolsep}{5pt} 
	\renewcommand{\arraystretch}{1.2} 
	
	\begin{table*}
		\centering
		\begin{tabular}{|l||r|r|r|r||r|r|r|r|}
			\hline 		  
			MTS & \multicolumn{4}{|c||}{discharge} &  \multicolumn{4}{|c|}{clinical}  \\ 	
			set 1 & \multicolumn{4}{|c||}{summaries} &  \multicolumn{4}{|c|}{notes} \\\hline
			scope 		& $\infty$  & 	  5 &     4 & 3		&  $\infty$ &     5 &     4 &	 3	\\ \hline  
			TP 			& 107   & 	105 & 	102 &  86 		& 	336 &	335 & 	333 & 	332 \\ 	\hline
			TN 			& 392   & 	403 & 	407 & 407 		& 	253 &	267 & 	271 & 	277 \\ \hline		
			FP 			&  17   & 	  6 &     2 &   2 		& 	 31 &	 17 & 	 13 & 	  7 \\ 	\hline
			FN 			&   1   & 	  3 & 	  6 &  22 		& 	  2 &	  3 & 	  5 & 	  6 \\ 	\hline\hline
			Acc			& 0.965 & 0.983 & {\bf 0.985} & {\em 0.954}		& {\em 0.947} & 0.968 & 0.971 & {\bf 0.979} \\ 	\hline
			Prec		& {\em 0.863} & 0.946 & {\bf 0.981} & 0.977		& {\em 0.916} & 0.952 & 0.962 & {\bf 0.979}	\\ 	\hline
			Rec			& {\bf 0.991} & 0.972 & 0.944 & {\em 0.796}		& {\bf 0.994} & 0.991 & 0.985 & {\em 0.982} \\ \hline		
			F1			& 0.922 & 0.959 & {\bf 0.962} & {\em 0.878}		& {\em 0.953} & 0.971 & 0.974 & {\bf 0.981}	\\ 	\hline		
			
		\end{tabular}\\ 
		\caption{Evaluation results for set 1 and trigger set MTS}
		\label{tab:compNegEx1MTS}
	\end{table*}
}

{
	\setlength{\tabcolsep}{5pt} 
	\renewcommand{\arraystretch}{1.2} 
	
	\begin{table*}
		\centering
		\begin{tabular}{|l||r|r|r|r||r|r|r|r|}
			\hline 		  
			OTS & \multicolumn{4}{|c||}{discharge} &  \multicolumn{4}{|c|}{clinical}  \\ 	
			set 1 & \multicolumn{4}{|c||}{summaries} &  \multicolumn{4}{|c|}{notes} \\\hline
			scope 		& $\infty$  & 	  5 &     4 & 3		&  $\infty$ &     5 &     4 &	 3	\\ \hline  
			TP 			& 105   & 	103 & 	97 &  79 		& 	334 &	334 & 	333 & 	332 \\ 	\hline
			TN 			& 391   & 	404 & 	406 & 406 		& 	253 &	267 & 	271 & 	277 \\ \hline		
			FP 			&  18   & 	  5 &     3 &   3 		& 	 31 &	 17 & 	 13 & 	  7 \\ 	\hline
			FN 			&   3   & 	  5 & 	  11 &  29 		& 	  3 &	  4 & 	  5 & 	  6 \\ 	\hline\hline
			Acc			& 0.959 & {\bf 0.981} & 0.973 & {\em 0.938}		& {\em 0.944} & 0.966 & 0.971 & {\bf 0.979} \\ 	\hline
			Prec		& {\em 0.854} & 0.954 & {\bf 0.970} & 0.963		& {\em 0.915} & 0.952 & 0.962 & {\bf 0.979}	\\ 	\hline
			Rec			& {\bf 0.972} & 0.954 & 0.898 & {\em 0.731}		& {\bf 0.988} & 0.988 & 0.985 & {\em 0.982} \\ \hline		
			F1			& 0.909 & {\bf 0.954} & 0.933 & {\em 0.832}		& {\em 0.950} & 0.970 & 0.974 & {\bf 0.981}	\\ 	\hline		
			
		\end{tabular}\\ 
		\caption{Evaluation results for set 1 and trigger set OTS}
		\label{tab:compNegEx1OTS}
	\end{table*}
}
{
	\setlength{\tabcolsep}{5pt} 
	\renewcommand{\arraystretch}{1.2} 
	
	\begin{table*}
		\centering
		\begin{tabular}{|l||r|r|r|r||r|r|r|r|}
			\hline 		  
			MTS & \multicolumn{4}{|c||}{discharge} &  \multicolumn{4}{|c|}{clinical}  \\ 	
			set 2 & \multicolumn{4}{|c||}{summaries} &  \multicolumn{4}{|c|}{notes} \\\hline
			scope 		& $\infty$  & 	  5 &     4 & 3			&  $\infty$ &     5 &     4 &	 3	\\ \hline  
			TP 			&  684   &  655 &	 613 &  555 		& 	302 &	302 & 	301 & 	300 \\ 	\hline
			TN 			& 1743   & 1792 &	1811 & 1821 		& 	220 &	246 & 	248 & 	254 \\ \hline		
			FP 			&  111   & 	 62 &     43 &   33 		& 	 41 &	 15 & 	 13 & 	  7 \\ 	\hline
			FN 			&   36   & 	 65 & 	 107 &  165 		& 	  4 &	  4 & 	  5 & 	  6 \\ 	\hline\hline
			Acc			& 0.943 & {\bf 0.951} & 0.942 & {\em 0.923}		& {\em 0.921} & 0.966 & 0.968 & {\bf 0.977} \\ 	\hline
			Prec		& {\em 0.860} & 0.914 & 0.934 & {\bf 0.944}		& {\em 0.880} & 0.953 & 0.959 & {\bf 0.977}	\\ 	\hline
			Rec			& {\bf 0.950} & 0.910 & 0.851 & {\em 0.771}		& {\bf 0.987} & 0.987 & 0.984 & {\em 0.980} \\ \hline		
			F1			& 0.900 & {\bf 0.902} & 0.891 & {\em 0.849}		& {\em 0.929} & 0.966 & 0.966 & {\bf 0.974}	\\ 	\hline		
			
		\end{tabular}\\ 
		\caption{Evaluation results for set 2 and trigger set MTS}
		\label{tab:compNegEx2MTS}
	\end{table*}
}

{
	\setlength{\tabcolsep}{5pt} 
	\renewcommand{\arraystretch}{1.2} 
	
	\begin{table*}
		\centering
		\begin{tabular}{|l||r|r|r|r||r|r|r|r|}
			\hline 		  
			OTS & \multicolumn{4}{|c||}{discharge} &  \multicolumn{4}{|c|}{clinical}  \\ 	
			set 2 & \multicolumn{4}{|c||}{summaries} &  \multicolumn{4}{|c|}{notes} \\\hline
			scope 		& $\infty$  & 	  5 &     4 & 3		&  $\infty$ &     5 &     4 &	 3	\\ \hline  
			TP 			&  692   & 	 654 & 	 629 &  565 		& 	302 &	302 & 	301 & 	300 \\ 	\hline
			TN 			& 1741   & 	1789 & 	1800 & 1816 		& 	222 &	247 & 	248 & 	254 \\ \hline		
			FP 			&  113   & 	  65 &    54 &   38 		& 	 39 &	 14 & 	 13 & 	  7 \\ 	\hline
			FN 			&   28   & 	  66 & 	  91 &  155 		& 	  4 &	  4 & 	  5 & 	  6 \\ 	\hline\hline
			Acc			& 0.945 & {\bf 0.949} & 0.944 & {\em 0.925}		& {\em 0.924} & 0.968 & 0.968 & {\bf 0.977} \\ 	\hline
			Prec		& {\em 0.860} & 0.910 & 0.921 & {\bf 0.937}		& {\em 0.886} & 0.956 & 0.959 & {\bf 0.977}	\\ 	\hline
			Rec			& {\bf 0.961} & 0.908 & 0.874 & {\em 0.785}		& {\bf 0.987} & 0.987 & 0.984 & {\em 0.980} \\ \hline		
			F1			& 0.909 & {\bf 0.909} & 0.897 & {\em 0.854}		& {\em 0.934} & 0.971 & 0.971 & {\bf 0.979}	\\ 	\hline		
			
		\end{tabular}\\ 
		\caption{Evaluation results for set 2 and trigger set OTS}
		\label{tab:compNegEx2OTS}
	\end{table*}
}

We also studied which triggers occurred most often. It turned out that looking at the notes the terms \textit{keine}, \textit{kein}, \textit{ohne}, and \textit{Ausschluss} (no, without, and exclusion) cover over 98\% of negations. 
The evaluation reveals some interesting aspects:
\begin{enumerate}
	\item The results on clinical notes are almost identical when comparing the two trigger sets on the same data. This is probably caused by the fact that within notes, negations are mostly expressed with simple negation terms.
	\item When comparing the results of different window sizes, there is a noticeable difference between the two types of documents: the best F1 values on notes can be achieved with window size 3, the worst with unlimited size. 	Almost the opposite is true for summaries: window size 3 performs worst, size 5 is best (except for MTS on set 1, where 4 is slightly better). This can be explained by the more complex phrasings in the summaries where more adjectives are used (more complex adverbial modifier constructions are used) that increase the distance between negations and negated terms.
	\item The difference between using MTS and  OTS triggers is small: on the first set for discharge summaries the best F1 value 	for MTS is 0.962, the best for OTS is 0.954, on the second set its 0.902 and 0.909.
	
	\item When examining the false positives of summaries (evaluation set 2), the following causes were most frequent: 1) faults due to the propagation of a trigger too far, 2) wrong assignment (according to the ground truth) when a term (probably) just negates an adjective as in ``kein sicherer Anhalt'' (no reliable indication)\footnote{These cases are ambiguous, as in ``no bacterial infection'', for example it is not clear if there is any other kind of infection.}. Another interesting case is ``Zeichen rechts positiv, links negativ'' (signs right positive, left negative), where a term seems both positive and negative and has to be discriminated or a complex term to be constructed first.

\end{enumerate}

Furthermore we observed that NegEx does not handle interferences between pre- and post-negation triggers: let's assume ``not'' is a pre-negation trigger and ``not visible'' a post-negation. Let's further regard the sentence ``lesions are not visible''. The implementations of NegEx and ConText both iterate over the words of a sentence and replace parts matching a trigger by special tags. As it uses pre-negations prior to post-negations in our example ``not'' will be replaced by ``$<$NEG\_PRE$>$''. In the resulting sentence ``lesions are $<$NEG\_PRE$>$ visible'' the post-negation trigger ``not visible'' is no longer present and the correct negation will therefore not be annotated. 
We adopted our own implementation to consider this special case by doing an extra post-check if a pre-negation did not cover any concept.

\subsection{Pipeline 2: Dependency parsing}

We originally planned to compare the results of two methods to detect negations: NegEx and dependency parsing. During our implementations and after some experiments we had to drop the idea of a quantitative evaluation of the dependency parser based on the following reasons:
1) phrasings and negation patterns in our example documents turned out to be mostly pretty simple (as shown above), therefore the NegEx algorithm produces very good results and there is pragmatically not much room for improvement, 2) applying the Stanford dependency parser to German medical texts turned out to be much more difficult and resource-intensive than expected. We describe the qualitative results in the following. 

The selected examples of our dataset used to demonstrate the potentials of dependency parsing are encouraging but they just don't occur often enough in our documents to be significant. For example, one possible source of incorrect results of NegEx are double and pseudo-negations. Phrases like ``x ist nicht vorhanden'' (x is not present) or  ``x ist nicht ausgeschlossen'' (x is not excluded) are very similar but the latter does not negate the existence of x but the \textit{exclusion} of x which says little about the actual existence of x. Therefore it's called a pseudo-negation.
As a consequence it is not sufficient to use ``nicht'' or  ``ausgeschlossen'' as triggers as both will produce a false positive result on this sentence. ``not excluded'' is a double negation which must be handled special.

\begin{figure}
	\centering
	\includegraphics[width=1.0\columnwidth]{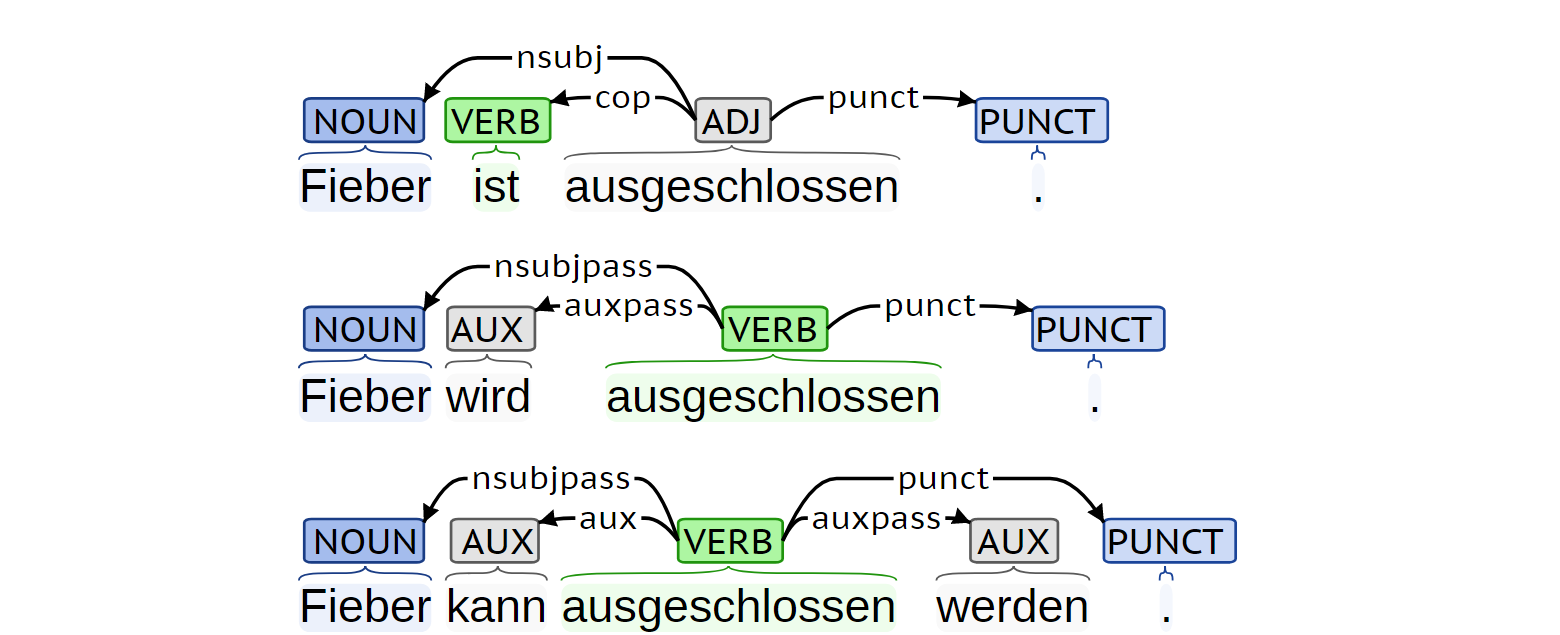}
	\cprotect\caption{Simple patterns covering several use cases, extended to except negated exclusions:
		\verb+{}</nsubj.*/ ({word:/ausgeschlossen/}+
		\verb+!>> /neg/ {})+}
	\label{fig:patterns1}
\end{figure}

\begin{figure}
	\centering
	\includegraphics[width=1.0\columnwidth]{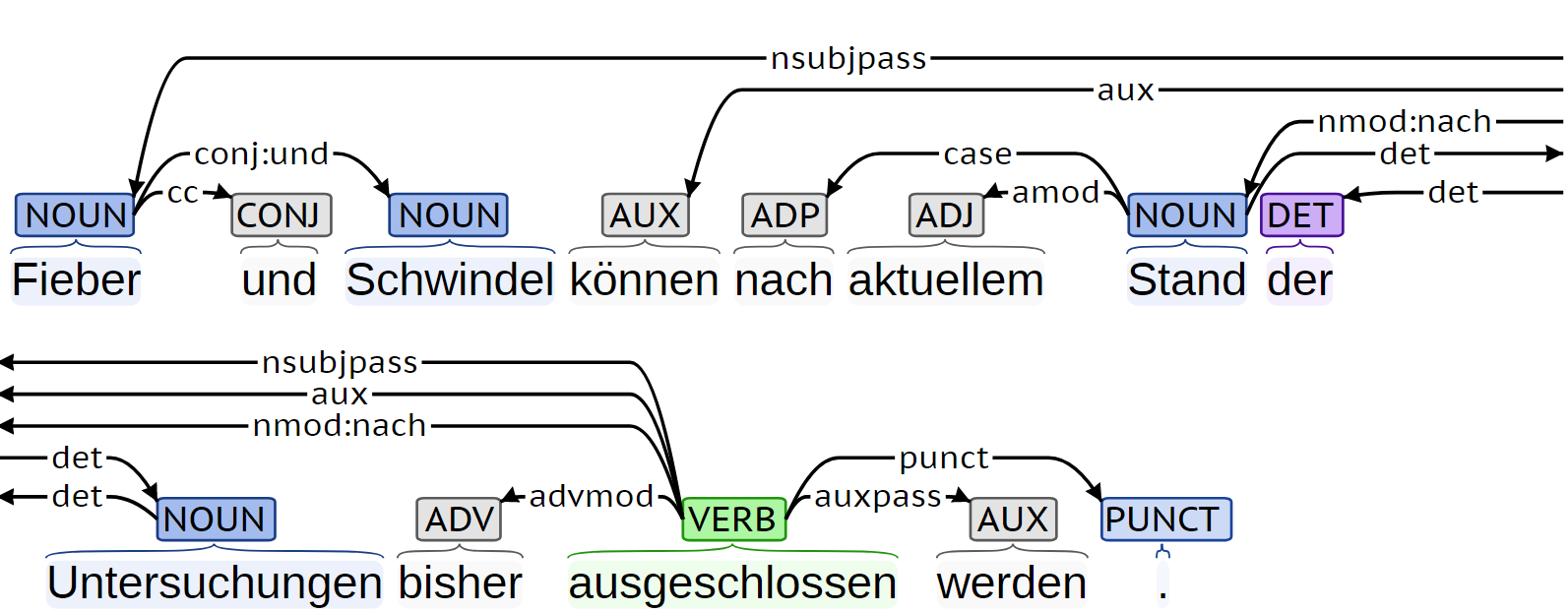}
	\cprotect\caption{Sentence with long inclusion}
	\label{fig:patterns11}
\end{figure}

To cover these cases we employed special rules on the dependency trees. 
We started by examining some simple sentences like in figure \ref{fig:patterns1}. The corresponding pattern matches sentences like ``Fieber ist ausgeschlossen.``, ``Fieber wird ausgeschlossen.'', and ``Fieber kann ausgeschlossen werden.'' (Fever is excluded. Fever can be excluded.).
This pattern even covers sentences where governor and dependents are far separated (in German the verb often appears at the end of a sentence), like
''\textit{Fieber und Schwindel k\"onnen nach aktuellem Stand der Untersuchungen bisher ausgeschlossen werden.}'' (see figure \ref{fig:patterns11}: 
fever and dizziness can be excluded so far according to the current state of research).
``\verb+!>> neg+'' is added to ensure negated verbs are \textit{not} matched 
(remember `$>>$' is a relation chain and the preceding `$!$' excludes such a chain). These long distance phrases can be covered when regarding the (semantic) dependency structure of a sentence.

\begin{figure}
	\centering
	\includegraphics[width=1.0\columnwidth]{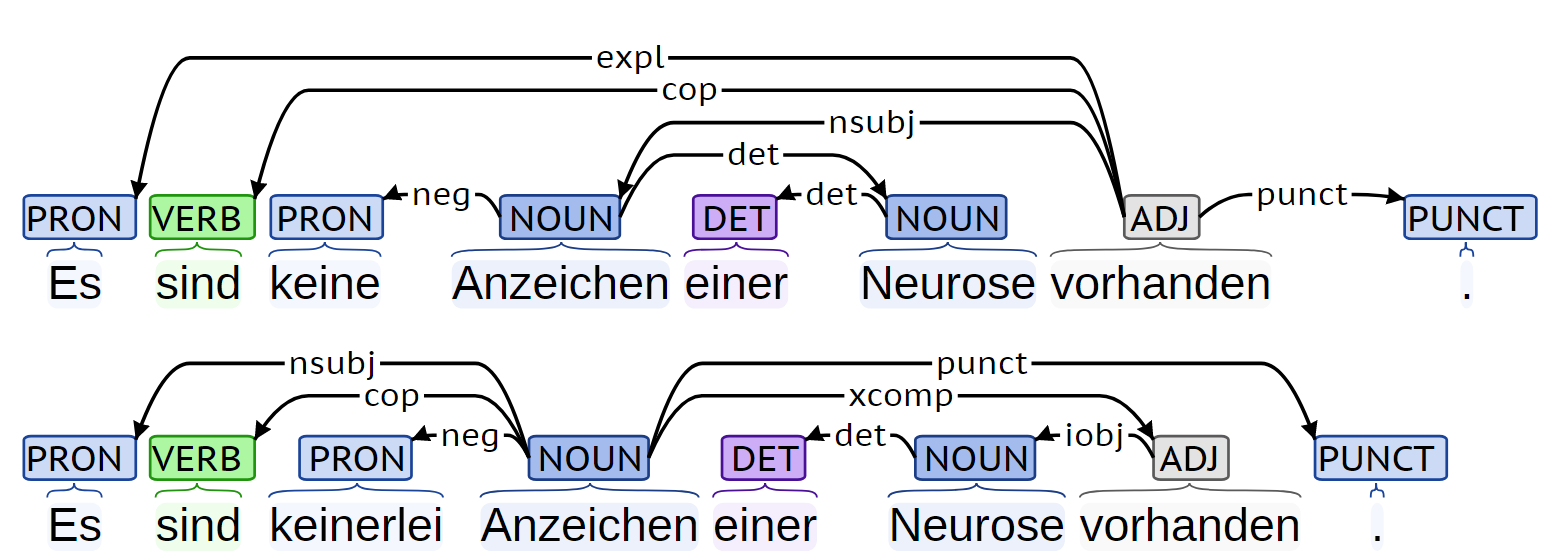}
	\cprotect\caption{Similar sentences with heavily differing parse trees.}
	\label{fig:keinerlei}
\end{figure}

On the other hand very similar sentences may produce different dependency parse trees and therefore are difficult to be covered with only a few patterns. Consider the  two sentences  of figure \ref{fig:keinerlei} that differ only in the word ``keine'' / ``keinerlei' (both words mean ``no'', ``none'', or ``not any'').

When you step deeper into specifying patterns for negations you soon encounter some challenges (at least in German). Sentences without a conjugated verb are not well covered by the Stanford NLP dependency parser model. They often occur in medical short texts like findings where the auxiliary verb ``be'' is left out, e.g., ``no edema detectable'' or  ``no fever''.

\begin{figure}
	\centering
	\includegraphics[width=1.0\columnwidth]{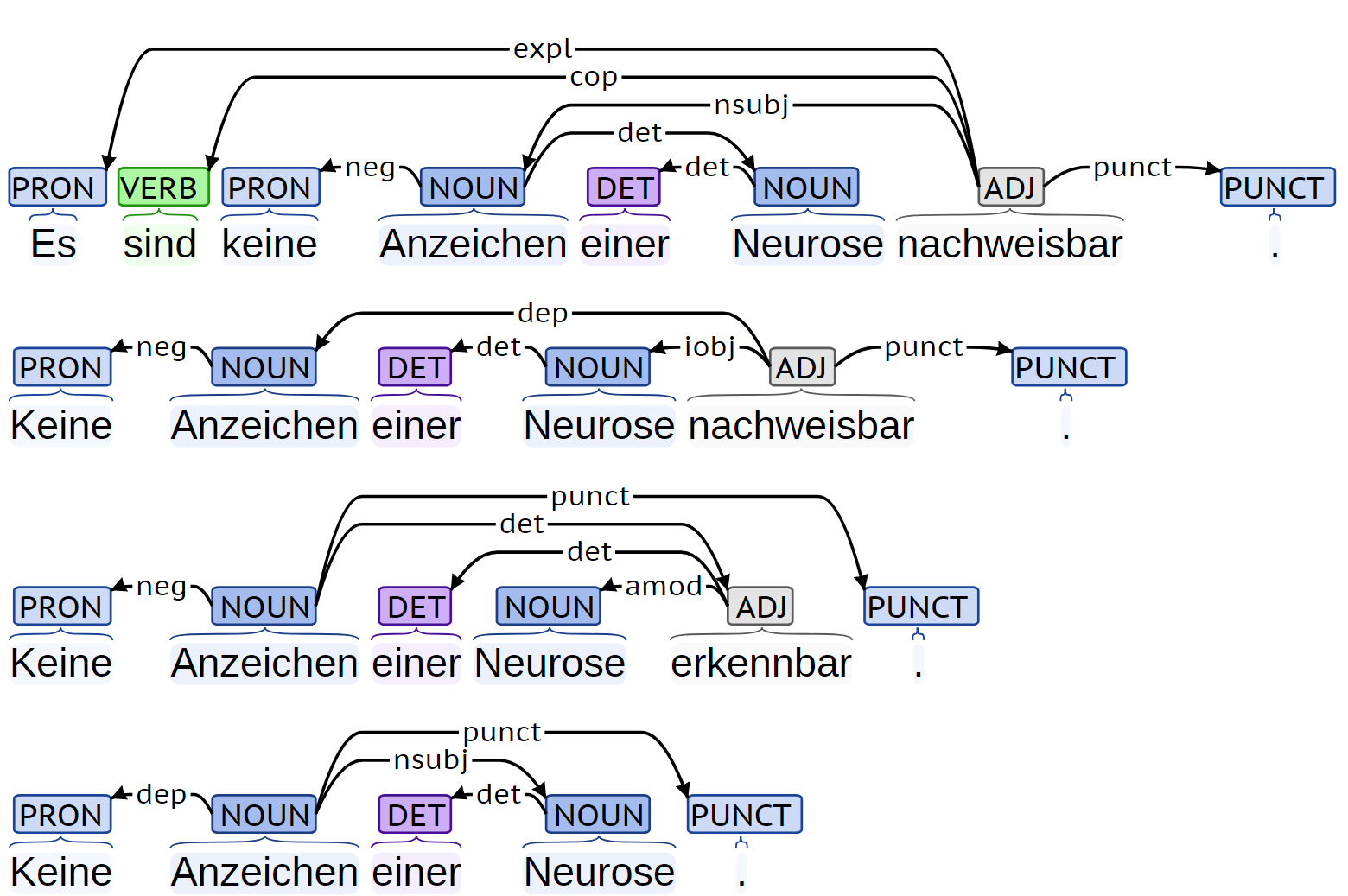}
	\cprotect\caption{Different semantic structures of similar phrases.}
	\label{fig:erkennbar}
\end{figure}

Figure \ref{fig:erkennbar} shows four very common wordings of the same fact (all stating that there are no signs of neurosis recognizable/verifiable). Although they differ only slightly,  their dependency parser outputs are very different and cannot be covered by the same pattern.

\begin{figure}
	\centering
	\includegraphics[width=1.0\columnwidth]{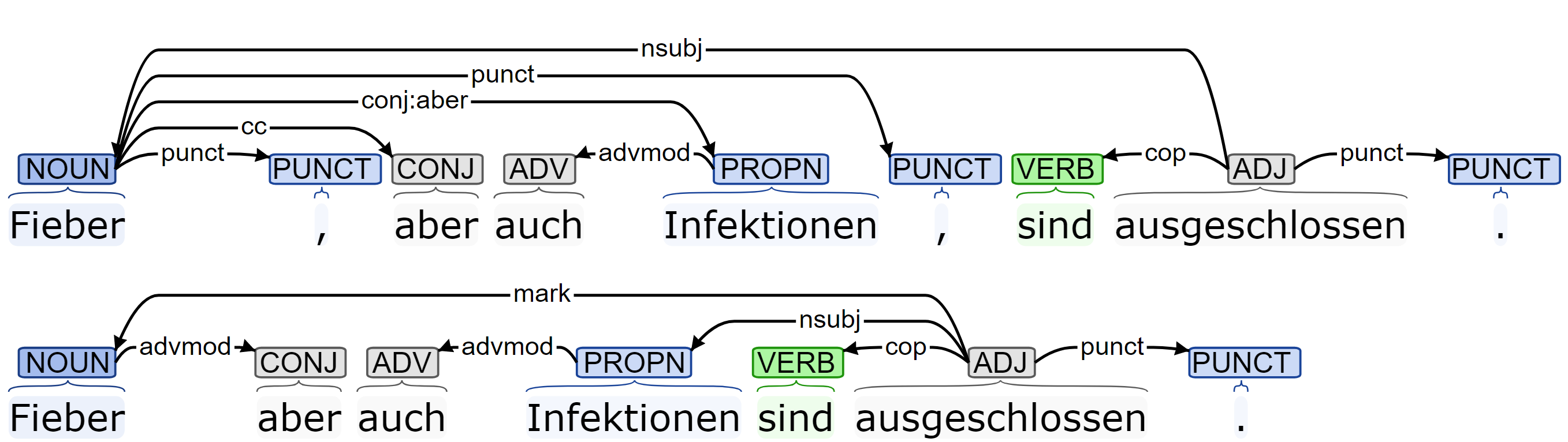}
	\cprotect\caption{Different semantic structures caused by punctuation.}
	\label{fig:punctuation}
\end{figure}

Incorrect or `sloppy' punctuation makes this effect even worse, as can be seen in figure \ref{fig:punctuation} on the sentence ``Fever as well as infections are excluded''.

\begin{figure}
	\centering
	\includegraphics[width=1.0\columnwidth]{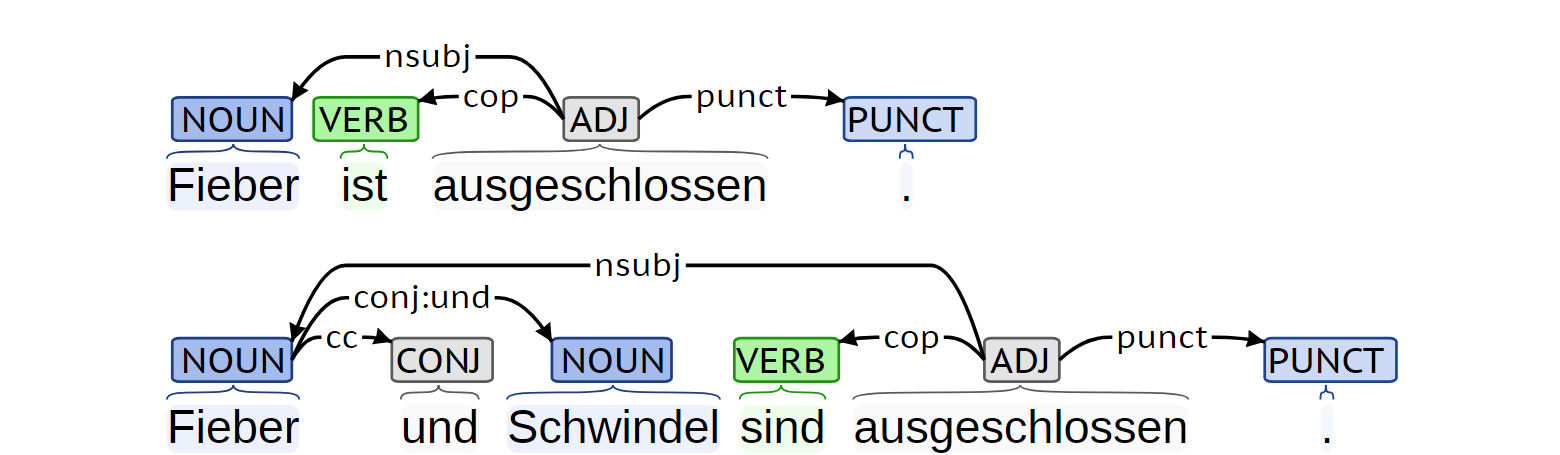}
	\cprotect\caption{Node conjunctions.}
	\label{fig:conjunction}
\end{figure}

On the other hand dependency parsing has the great advantage that patterns over dependency trees offer easy ways to \textit{exclude} specific wordings which is relevant to the mentioned double negations. Consider a simple negation pattern like ``kein x (no x)'', which can be covered by the pattern ``\verb|{} > /neg/ {}|''.
To exclude sentences like ``kein Ausschluss von x'' (no exclusion of x) we can extend the pattern to 
\verb|!{lemma:/ausschluss/}=dep >/neg/{}|, which specifies that the  dependent of the negation must not be the word ``ausschluss'' (exclusion).

The need for a more detailed distinction of wordings becomes obvious if you look at the phrases ``Test auf Masern'' and ``Verdacht auf Masern'' (test/suspicion on measles). Although measles are mentioned in the first case this is no statement on their appearance. This could be flagged as a pseudo-positive wording. The second phrase at least increases the likelihood of measles and could be marked as a (positive) speculation. If we extend the phrase to ``Test auf Masern war positiv'' (test on measles was positive) measles have to be marked as affirmed.

\begin{figure}
	\centering
	\includegraphics[width=1.0\columnwidth]{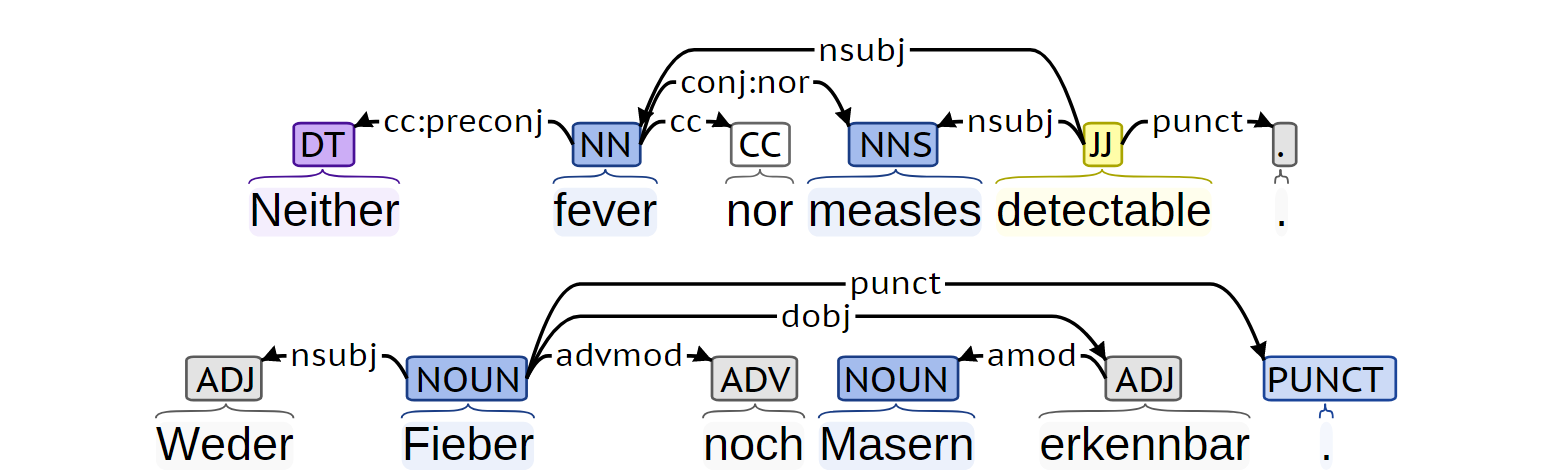}
	\caption{English: ``Neither x nor y'', German: ``Weder x noch y''}
	\label{fig:weder_noch}
\end{figure}

Our example data contained an interesting sentence: ``Weder enzymkinetisch noch elektrokardiografisch gab es Anhalt f\"ur eine kardiale Ischaemie'' (Neither enzyme kinetics nor electrocardiography gave evidence for cardiac ischemia). This form of surrounding trigger  ``weder ... noch ...'' (neither ... nor ...) cannot be covered by the NegEx patterns.
In contrast to NegEx, Semgrex patterns are more flexible and allow for the specification of such cases. However, the Stanford NLP model for German does not seem to know about this construction as a comparison of the resulting trees in English and German shows (see figure \ref{fig:weder_noch}). 

\section{General Results}

For our example documents a rather short list of triggers combined with a scope window size of 5 was sufficient to achieve good results with the NegEx algorithm. This set can easily be maintained and requires no special know-how. When regarding the ground truth, we reached a F1 value of at least 0.954.
Extending the set of triggers (like it was done by Cotik et al. \cite{DBLP:conf/coling/CotikRXUBS16}) from ``kein'' to ``Patient zeigt kein'', for example,  does not have any advantages. 

Our results in dependency parsing using the Stanford NLP parser depend very much  on the German model used by the parser, which--as we showed above--does not always seem to be well trained on wordings often used in medical texts. 

The examples above demonstrate the inherent difficulties of dependency parsing, namely the noticeably differences of parse trees, e.g.,
\begin{itemize}
	\item on omitting terms like `is' or `are' or a declined verb,
	\item on variants in punctuation,
	\item on using synonyms,
	\item slight changes in the sentence structure.
\end{itemize}

As these wording aspects are very common in medical texts (at least in documents of the type we tested on), handling all resulting parse trees is very cumbersome and tedious.  The correct coverage of more complex phrasings is of little importance in our case as they seldom occur in our text (if at all). Therefore the disadvantages of dependency parsing predominate significantly. The rather poor Stanford model for German adds an additional negative aspect.

Additionally the negations of most of our test phrases from medical texts can be easily covered by NegEx. The idea of combining both methods (NegEx and Dependency Parsing) and exploiting an additional analysis by dependency parsing failed due to the lacking robustness to small changes in surface structure.

\section{Discussion}

Structural information as provided by dependency parsers is useful for a variety of text types in the medical domain. Although \cite{DBLP:journals/bioinformatics/FundelKZ07, DBLP:conf/naacl/WilsonWH05, DBLP:conf/naacl/FinkelM09, DBLP:conf/bionlp/FinkelDNNMS04} reported on several success stories in bioinformatics, e.g., recognizing context polarity (e.g., negations of medical terms), named entity recognition, or the extraction of medical entities in medical web documents, the (structure and) results of the type of  examination reports as under investigation here is very different.

In order to evaluate and predict parser usefulness for negation detection, we reported on the accuracy  when compared to simple string parsing methods on the surface form. Evaluating the parsers using task specific metrics, we have shown that dependency parsing does not lead to a better average accuracy.

Dependency parsers provide information useful for a variety of down the line applications in medical texts, \cite{DBLP:conf/IEEEicci/ChenBR10} for example report on automatically assigning ICD-9-CM codes to patient records which is structurally similar to our extraction task. However, negation detection as found in our text repository cannot benefit from the advantages reported in previous works (\cite{DBLP:journals/bioinformatics/FundelKZ07, DBLP:conf/naacl/WilsonWH05, DBLP:conf/naacl/FinkelM09, DBLP:conf/bionlp/FinkelDNNMS04, DBLP:conf/IEEEicci/ChenBR10}). It seems that more sophisticated extraction tasks only (e.g., the automatic extraction of relations between medical concepts (\cite{DBLP:journals/jamia/RinkHR11})) which exhibit free-word order constructs or other complex syntactical constructions are needed. These can be characterized as information extraction tasks best described in terms of a set of binary relations that hold between the relevant constituents in more complex relations in a sentence of a medical text, an assumption which does not hold in our corpus. 

\begin{figure}
	\centering
	\includegraphics[width=1.0\columnwidth]{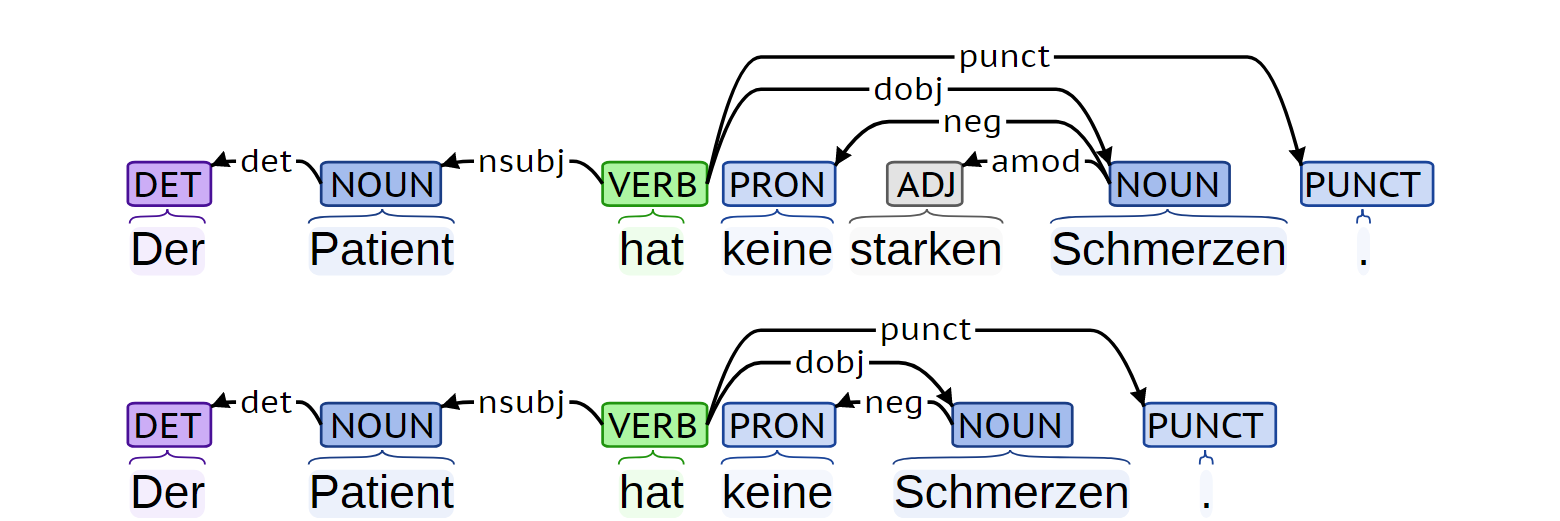}
	\caption{Unclear association of negations.}
	\label{fig:unclear}
\end{figure}

Sometimes the reference of a negation in a sentence is ambiguous.
Although it is argued (e.g., in \cite{DBLP:conf/coling/CotikRXUBS16}) that in these cases dependency parsing would help, the following example shows that this is not always the case: It is not clear whether ``Keine starken Schmerzen'' (no strong pain) should be interpreted as ``no pain at all'' or just ``no \textit{strong} pain''. The Stanford dependency parser does not change the assertion of the negation to the noun, see figure \ref{fig:unclear}, which is the same as we get from  NegEx.

\section{Conclusion}

We presented our work on detecting negations of concepts in medical texts like discharge summaries and clinical notes. We integrated a variant of the NegEx algorithm in our information extraction framework and examined different configurations. As an alternative approach and to overcome shortcomings of NegEx, components for deeper natural language analysis were added.
Both methods showed some advantages as well as drawbacks. 

NegEx is an easy-to-implement, simple algorithm based on a list of triggers easily maintained and fairly clear. It allows for a simple handling of triggers, but relies very much on the surface structure of sentences and a complete covering of variants, which can be rather cumbersome in German. The vague definition of the scope of negation triggers and the associated inability to handle long distance phrases are its main disadvantages. Some wordings like ``weder ... noch'' (neither ... nor) that enclose a term cannot be expressed.

Applying patterns on the result of dependency parsing appears to overcome the disadvantages of triggers depending on regular expressions (like in NegEx). Semgrex patterns can match long distance phrases without any additional effort by the programmer, conjuncted terms are automatically included,  enclosing patterns may cover additional cases. On the other hand the patterns are much harder to define and understand (requires  expertise in linguistics) and the handling of a larger list of patterns is rather cumbersome. Every simple and small word changes in a sentence may unpredictably alter the corresponding (variant of the) dependency structure  leading to the need to check every pattern when considering new phrases and every example sentence when adding new patterns. Our results show that dependency parsing is neither a good alternative nor a satisfactory complement, at least for our type of German medical documents with the given internal dependency parsing mechanism; this is counter-intuitive to the fact that head-dependent relations should provide an approximation to the negation scope and this should make them useful for medical IE tasks.

Based on our empirical evaluation, we would propose the following procedure:
\begin{itemize}
	\item start with a simple implementation of NegEx,
	\item check the types of negations in your documents,
	\item define corresponding simple triggers reduced to a minimum of words,
	\item evaluate NegEx with these triggers and different scope values,
	\item if complex negation phrases are common to your texts, you may try to cover them with dependency parsing.
\end{itemize}

\section*{Acknowledgements}
This research is part of the project ``clinical data intelligence'' (KDI) which was founded by the Federal Ministry for Economic Affairs and Energy (BMWi) and part of the new pAItient project founded by the Federal Ministry of Health (BMG). 

\bibliography{casestudy}

\end{document}